\documentclass[sigconf]{acmart}

\usepackage{booktabs} 

\usepackage{url}
\usepackage{epsfig}
\usepackage{bm}
\usepackage{balance}
\usepackage{enumitem}

\newcounter{inlineenum}
\renewcommand{\theinlineenum}{\arabic{inlineenum}}

\setcopyright{none}

\begin{document}
\fancyhead{}

\title{DCH-2: A Parallel Customer-Helpdesk Dialogue Corpus with Distributions of Annotators' Labels}

\author{Zhaohao Zeng}
\affiliation{%
  \institution{Waseda University, Tokyo, Japan}
}
\email{zhaohao@fuji.waseda.jp}

\author{Tetsuya Sakai}
\affiliation{%
  \institution{Waseda University, Tokyo, Japan}
}
\email{tetsuyasakai@acm.org}


\begin{abstract}
We introduce a data set called DCH-2,
which contains 4,390 real
customer-helpdesk dialogues in Chinese
and their English translations.
DCH-2 also contains
dialogue-level annotations 
and turn-level annotations
obtained independently from either 19 or 20 annotators.
The data set was built through 
our effort as organisers of the 
NTCIR-14 Short Text Conversation
and NTCIR-15 Dialogue Evaluation tasks,
to help researchers understand
what constitutes an effective customer-helpdesk dialogue,
and thereby 
build efficient and helpful helpdesk systems that are available
to customers at all times.
In addition,
DCH-2 may be utilised for other purposes,
for example, as a repository for
retrieval-based dialogue systems,
or as a parallel corpus
for machine translation in the helpdesk domain.
\end{abstract}




\keywords{
dialogue;
evaluation;
helpdesk;
corpus;
}

\maketitle


\section{Introduction}\label{s:intro}

Companies have call centers and helpdesks for their products and services
to help their customers whenever they encounter problems.
If the helpdesk staff can be replaced completely by intelligent dialogue systems,
this would mean a lot of cost saving from the companies' viewpoint.
More importantly, this would enable
a 24-hour, wait-free, efficient and homogeneous helpdesk service for the customers.
However, existing natural language dialogue systems are 
not effective enough to completely replace humans.
Effective helpdesk systems should understand the customer's problem,
gather the necessary context, and provide a solution to the problem
through minimal interactions.

To help researchers build effective dialogue systems for helpdesk applications,
we have constructed a Chinese-English parallel corpus of customer-helpdesk dialogues,
with dialogue-level and turn-level annotations.
The dialogue-level annotations are given as distributions of dialogue quality scores
as rated by multiple annotators,
while the turn-level annotations
are given as distributions over \emph{nugget types}:
that is, 
whether a turn states the problem the customer is facing,
whether it provides useful information for solving it, 
or 
whether it shows that the problem has been solved (details to follow in Section~\ref{ss:nugget-annotation}).

Our data set, \emph{DCH-2} (Dialogues between a Customer and a Helpdesk - Version 2)
was constructed through our effort as the organisers 
of the NTCIR-14 Short Text Conversation (STC-3)~\cite{zeng19} and
NTCIR-15 Dialogue Evaluation (DialEval-1)~\cite{zeng20} tasks.
These two tasks offered the same subtasks: 
the \emph{Dialogue Quality} (DQ) subtask
and the \emph{Nugget Detection} (ND) subtask.
In the DQ subtask, participants were 
required to estimate, for each test dialogue,
the gold dialogue quality score distributions;
in the ND subtask, participants were 
required to estimate, for each turn in the dialogue,
the gold distribution over nugget types.
The DCH-2 dialogue corpus contains the 
DCH-1 corpus (training data used at STC-3),
the STC-3 test dialogues (additional training data used at DialEval-1),
and the DialEval-1 test dialogues.
However, DCH-2 may also be useful for researchers 
working on tasks other than the DQ and ND tasks.
For example, retrieval-based helpdesk response systems
may find our data useful as a knowledge repository;
Chinese-English
machine translation systems may leverage DCH-2 as 
a parallel corpus for the helpdesk domain.
\begin{figure*}[!htb]
\begin{center}
\includegraphics[width=1\textwidth]{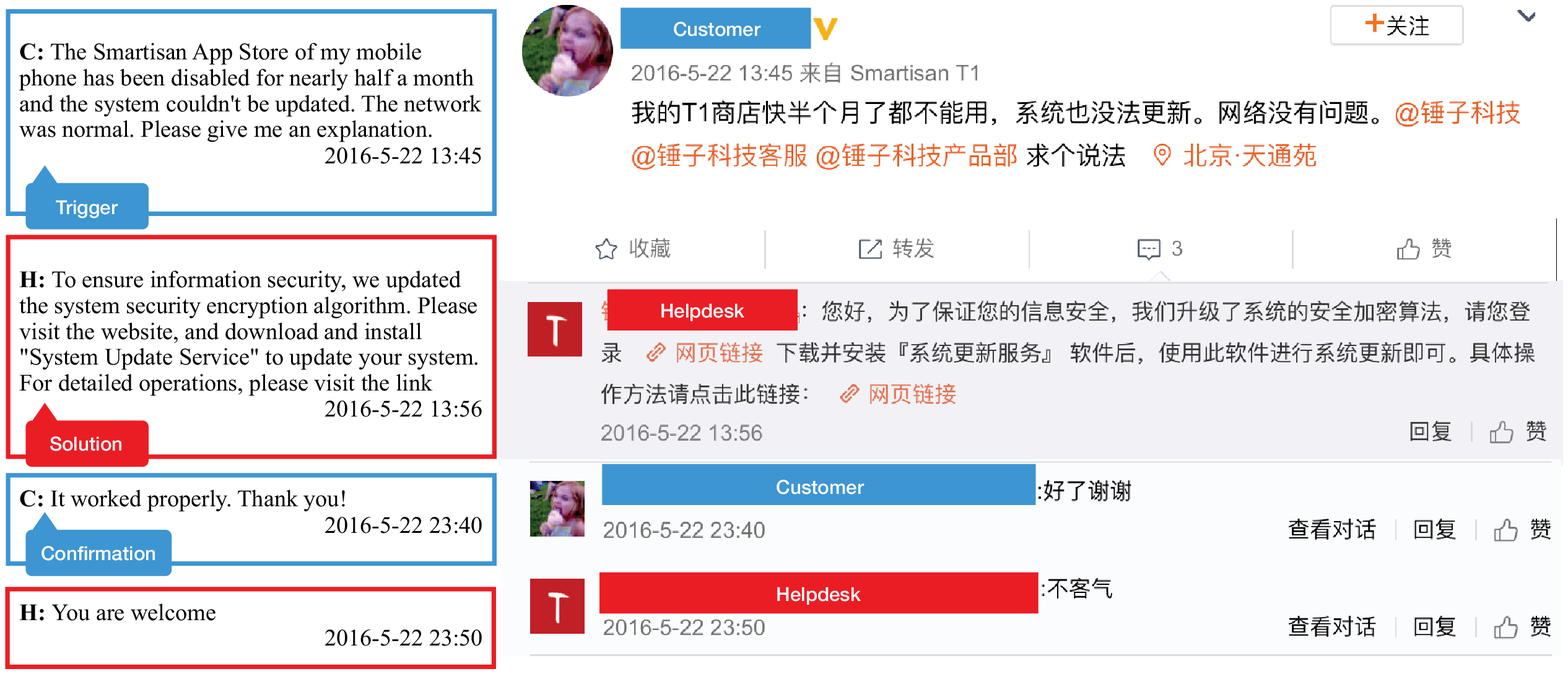}
\caption{An example of a dialogue between
a customer (C) and a helpdesk (H). The left part is the translated dialogue and the right part is the screenshot of the original dialogue on Weibo.}
\label{f:dialogue-example}
\end{center}
\end{figure*}
\section{Related Work}\label{s:related}

The DARPA COMMUNICATOR Program that evaluated spoken dialogue systems in the travel planning domain~\cite{walker01} produced the Communicator 2000 Corpus, which consists of 662 dialogues based on nine different systems,
with per-call survey results on dialogue efficiency, dialogue quality, task success and user satisfaction. Unlike DCH-2,
these task-oriented dialogues focused on a {\em structured} task, and they evaluated these dialogues based on how they filled the key slots (e.g., ``origin,'' ``destination,'' ``date'') .

Lowe {\em et al.}~\cite{lowe15} released the Ubuntu Dialogue Corpus,
which contains 930,000 {\em human-human} dialogues extracted from Ubuntu chats.
They focus primarily on {\em unstructured} dialogues like we did in DCH-2. {However, while they automatically disentangled the chats to form dyadic dialogues, their original chat logs usually involve more than two parties, which makes it different from our dyadic customer-helpdesk DCH-2 dataset. 
}
They formed a response selection test data set by setting aside 2\% of the corpus
and forming ({\em context}, {\em response}, {\em flag}) triplets based on this set.
Here, context is the sequence of utterances that appear prior to the response
in the dialogue; response is either the actual correct response from the dialogue
or a randomly chosen utterance from outside the dialogue (but within the test set);
{\em flag} is one for the correct response and zero for incorrect responses.
For each correct response, they generated nine additional triplets containing different 
incorrect responses.
Thus, response selection systems are given a context and ten choices of responses,
and required to select one or more responses.
They use recall at $k$ as the evaluation measure, 
where $k$ is the size of the set of responses selected by the system
and therefore ``recall at 1'' reduces to accuracy.
Note that this evaluation setting does not require annotations for defining the gold standard.
They do not consider {\em ranked} lists of responses.

One novelty of DCH-2 is that
both the gold DQ and gold ND data 
are provided as distributions of labels 
rather than a single consolidated labels.
A gold DQ distribution is defined over 
an ordinal scale (namely, Likert scale scores): 
the task of estimating such a distribution 
is known as the \emph{ordinal quantification} task~\cite{SemEval2016task4,SemEval2017task4,SemEval2018Task1}.
In contrast, a gold ND distribution is defined over
nominal classes (namely, nugget types):
hence, the task of estimating this distribution
should probably be called \emph{nominal quantification}.
The reason why we preserve the distribution of labels given by multiple annotators
is that these assessments are inherently subjective.
If a user of DCH-2 prefers 
to evaluate DQ and ND tasks
in a deterministic way,
they can easily take the most frequent class from each of our gold distribution
and treat it as the gold class
for ordinal and nominal \emph{classification} tasks.

The Dialogue Breakdown Detection Challenge (DBDC) data sets~\cite{higashinaka19,YuikoTsunomori2020DSI-G}
also contain turn-level annotations that can be used
for evaluating ordinal quantification.
The dialogues are chats between a system and human (in English and in Japanese),
and each system turn is labelled 
with a distribution over
three ordinal classes,
namely,
\emph{breakdown}, \emph{possible breakdown}, and \emph{not a breakdown}.
While the organisers of DBDC used \emph{nominal} quantification measures
(namely, Jensen-Shannon Divergence and Mean Squared Error)
despite the ordinal nature of their classes, 
subsequently 
they examined nominal quantification measures used in the aforementioned NTCIR tasks,
namely, Normalised Match Distance (NMD) and Root Symmetric Normalised Order-aware Distance (RSNOD)~\cite{sakai18sigir}.
They reported that RSNOD is suitable for evaluating DBDC systems.
Section~\ref{s:tool} discusses a tool we have released for computing these measures.

\section{Dialogues}
DCH-2 is a data collection based on {\em real}  (i.e., human-human) customer-helpdesk dialogues. We crawl Weibo, a Chinese microblog service, to obtain customer-helpdesk dialogues, because many technology company accounts handle users' inquires on Weibo publicly. 

The dialogues in DCH-2 were crawled in three batches, with the same approach. The first batch (3700 dialogues) was crawled in 2016, and these dialogues formed a dialogue corpus namely DCH-1. The second batch (390 dialogues) crawled in 2018 were used as the test collection in NTCIR-14 STC-3 NDDQ subtasks~\cite{zeng19}. The third batch (300 dialogues) was crawled in 2020, and was used as the test collection of NTCIR-15 DialEval-1 Task~\cite{zeng20}.
\begin{table*}[htb]
\begin{center}

\caption{DCH-1 and DCH-2 dialogue corpus statistics.}\label{t:d-statistics}

\begin{tabular}{l|c|c}
\hline
					&DCH-1~\cite{zeng18}		&DCH-2\\
\hline
Data timestamps		& Jan. 2013 - Sep. 2016	&Jan. 2013 - Sep. 2016 (DCH-1 corpus)\\
					&						& Oct. 2016 - Apr. 2018  (NTCIR-14 STC-3 NDDQ test dialogues~\cite{zeng19})\\
					&						& Apr. 2018 - Jul. 2019 (NTCIR-15 DialEval-1 test dialogues~\cite{zeng20})\\
\hline
\#Chinese dialogues	&3,700					&4,390 (DCH-1 $+$ 390 STC-3 $+$ 300 DialEval-1)\\
\#English translations 	&1,264 (34\%)				&4,390 (100\%)\\
\#Helpdesk accounts	&161					&161 \\ 
Avg. \#turns/dialogue	&4.162					&4.201\\
Avg. turn length	 (\#chars)	&48.31					&54.541\\
\# Annotators per dialogue & 19 & 20\\
\hline
\end{tabular}

\end{center}
\end{table*}
\begin{figure*}[!htb]
\begin{center}
\includegraphics[width=.9\textwidth]{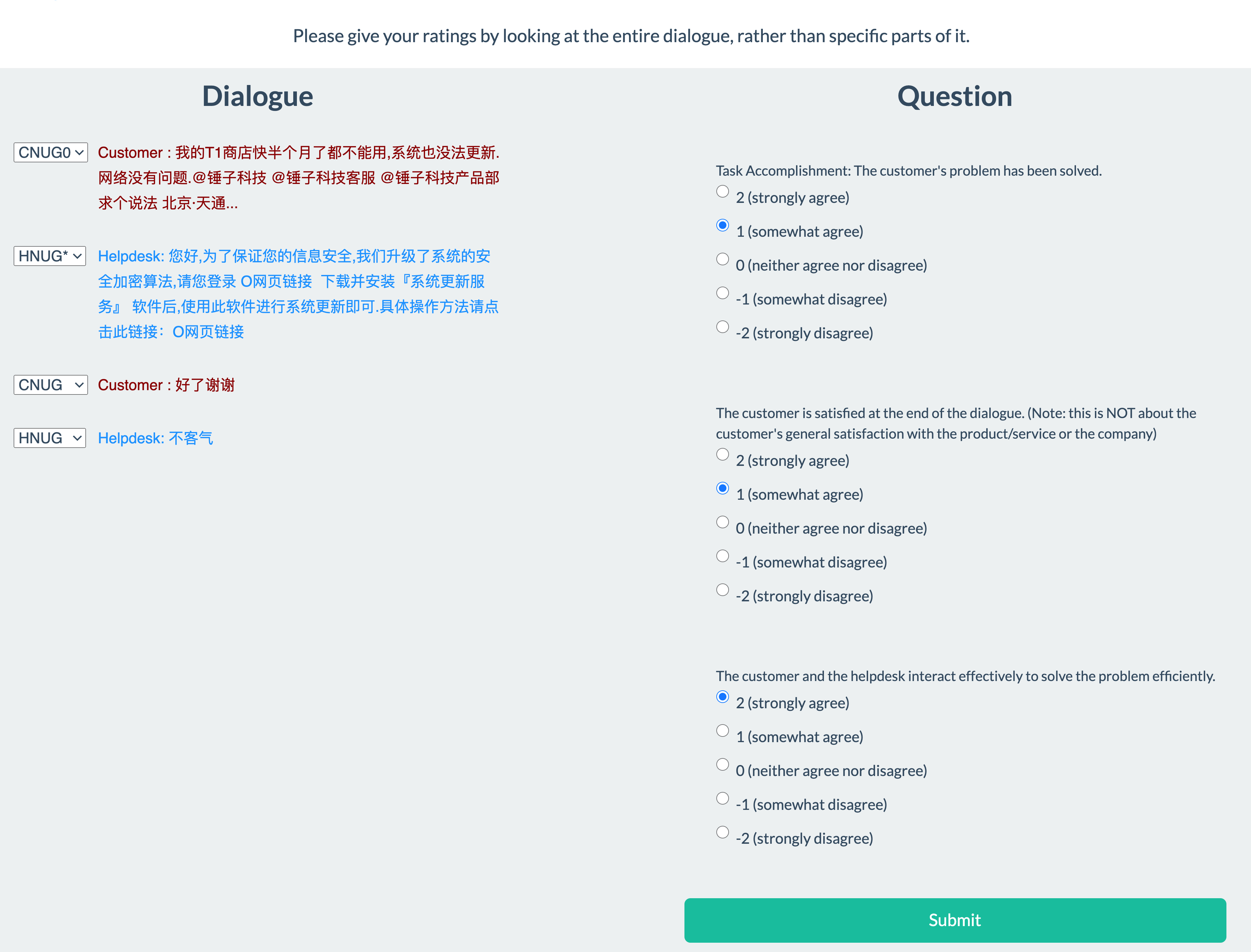}
\caption{The interface of the annotation tool. The Chinese dialogue on the left is the same as Figure~\ref{f:dialogue-example}.}
\label{f:interface}
\end{center}
\end{figure*}

\subsection{Dialogue Crawling}
We detail the mining process of the first batch (3,700 dialogues) as follows.

\begin{enumerate}

\item We collected an initial set of Weibo accounts by searching Weibo account names that contained keywords such as ``assistant'' and ``helper'' (in Chinese).
We denote this set by $A_{0}$. 
\item For each account name $a$ in $A_{0}$, 
we added a prefix ``@'' to $a$ and used the string as a query for 
searching up to 40 conversational threads (i.e., initial post plus comments on it) that contain a mention of the official account\footnote{
Weibo's interface for conversational threads is somewhat different from Twitter's:
comments to a post are not displayed on the main timeline; they are displayed under each post only if the ``comments'' button is clicked.
}. We then filtered out accounts that did not  respond to over one half of these threads.
As a result, we obtained 41 active account names.
We denote the filtered set of  ``active'' accounts as $A$.
\item For each account $a$ in $A$, we retrieved all threads that contain a mention of $a$ from January 2013 to September 2016, and extracted Customer-Helpdesk dyadic dialogues from them. We then kept those that consist of
at least one turn by Customer {\em and} one by Helpdesk.
As a result, 21,669 dialogues were obtained. This collection is denoted as $D_0$. 
Note that although we used account names in $A$ as seeds for searching the dialogue corpus,
we obtained dialogues involving not only these accounts but also \emph{subaccounts} of these accounts.
For example, when the customer mentions ``ABCD Company Helpdesk,''
a subaccount called ``ABCD Company Helpdesk Beijing'' might actually respond to it.
Such dialogues are also included in DCH-2;
thus it actually covers helpdesk accounts that are outside $A$. 
\item As $D_{0}$ is too large for annotation,
we sampled 3,700 dialogues from it based on the dialogue lengths.
For $i=2, 3, \ldots, 6$, we randomly sampled 700 dialogues that contained $i$ turns.
In addition, we randomly sampled 200 that contained $i=7$ turns;
we could not sample 700 dialogues for $i=7$ as $D_{0}$ did not contain enough dialogues that are very long.
\item To remove the privacy information in the dialogues, we replaced telephone numbers with 123456789, and replaced email addresses with XXX@YYY.com.
\end{enumerate}

When crawling the second batch (390 dialogues) and the third batch (300 dialogues), we utilise the same account names in $A$ as keywords to crawl from Weibo using the same crawling approach. However, we use different timestamps to filter the search results in different batches. In the first batch, we only crawl dialogues with the timestamps between January 2013 to September 2016. The timestamps of the second batch are between October 2016 to August 2018, and the timestamps of the third batch are between September 2018 to August 2020.

We utilised a similar method as the step (4) to sample dialogues based on the number of turns for the second batch (390 dialogues) and the third batch (300 dialogues). The second batch has 65 dialogues that contained 7 turns, and the third batch has 30 ones. For $i=2,3,......,6$, there are the same amount of the dialogues that contained $i$ turns.

\subsection{English Translations}

We hired professional translators to manually translate all the original Chinese dialogues into English.
Hence DCH-2 is a completely parallel corpus.
As we shall describe in Section~\ref{s:annotations},
the dialogue-level and turn-level annotations were performed solely
based on the Chinese part of the corpus;
since DCH-2 is a parallel corpus,
we assume that 
these annotations perfectly reflect the nature 
of the translated English dialogues as well.

Figure~\ref{f:dialogue-example} shows an actual dialogue from DCH-2, with its English version.

\subsection{DCH-2 Dialogue Corpus Statistics}
Table~\ref{t:d-statistics} provides the statistics of the dialogues in DCH-2, 
together with those of its predecessor, DCH-1.

\section{Annotations}\label{s:annotations}

This section describes how we obtained 
the dialogue-level and turn-level annotations.
The former was used as the gold data for the DQ subtask 
of NTCIR-14 STC-3 and NTCIR-15 DialEval-1;
the latter was used as the gold data for the ND subtask
of the above NTCIR tasks.

\subsection{Annotators}
We followed the same approach to annotate all the dialogues in DCH-2, but the dialogues were annotated in two groups. In 2018, we annotated the first 4,090 of 4390 dialogues with 19 annotators\footnote{We hired 20 annotators, but one did not complete the job. Thus, we ended up with 19 annotators.}. Note that DCH-1 was annotated by three annotators in a different way in 2016 \cite{zeng18}, but the 2016 annotations are not related to DCH-2. We revised the annotation criteria and annotated DCH-1 again in 2018 with the approach introduced in this section.  In 2019, we annotated the rest 300 dialogues with 20 annotators. We hired Chinese undergraduate students and graduate students
from Waseda University as annotators. Each annotator was asked to annotate all the dialogues in each group, so each dialogues was annotated by either 19 or 20 annotators. 
An initial face-to-face instruction and training session for the annotators was organised 
by the first author of this paper at Waseda University.
Each annotator was paid about 1,200 Japanese Yen per hour. We developed a web-based annotation tool for the annotation. Subsequently, the annotators were allowed to do their annotation work online
using a web-browser-based tool at their convenient location and time. Figure \ref{f:dialogue-example} shows a screenshot of the web-based annotation tool. The order of dialogues to annotators was randomised;
given a dialogue, each annotator was instructed to first 
read the entire dialogue carefully, and then complete the  
dialogue quality annotation criteria described 
in Section~\ref{ss:subjective-annotation};
finally, the annotators classified the nugget types for each dialogue turn,
where nugget types were defined as described in Section~\ref{ss:nugget-annotation}.
\begin{figure*}[!htb]
\begin{center}
\includegraphics[width=.8\textwidth]{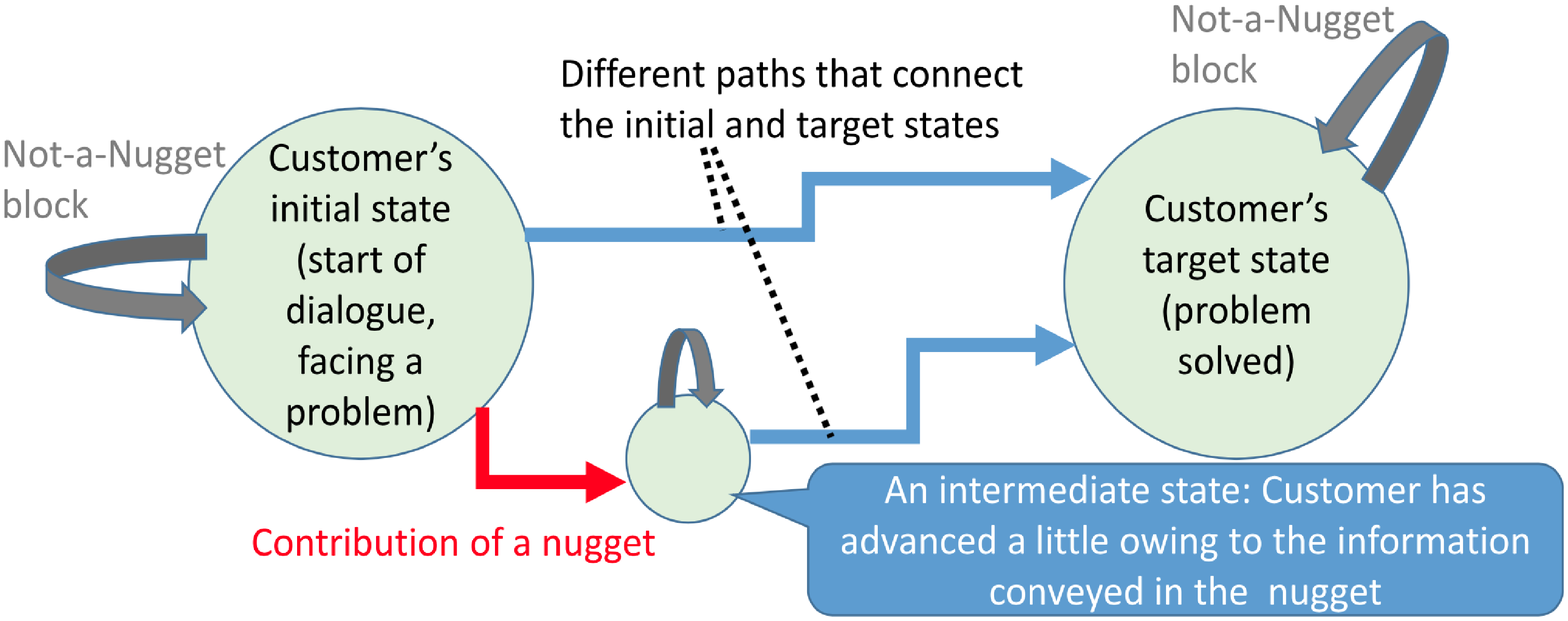}
\caption{Task accomplishment as state transitions, and the role of a nugget.}
\label{f:states}
\end{center}
\end{figure*}


\subsection{Dialogue Quality Annotations}\label{ss:subjective-annotation}
By Dialogue Quality (DQ) annotation, we mean manual quantification of the quality of a dialogue as a whole. Specifically, we introduce the following three dialogue quality scores for three different criteria.

\begin{description}
    \item [A-Score]: Task \textbf{A}ccomplishment (The customer's problem has been solved) 
    \item [S-score]: Customer \textbf{S}atisfaction (The customer is satisfied at the end of this dialogue.)
    \item [E-score]: Dialogue \textbf{E}ffectiveness (The customer and the helpdesk interact effectively to solve the problem efficiently.)
\end{description}

The annotators were asked to choose from the following options: 

2 (strongly agree), 1 (somewhat agree), 0 (neither agree nor disagree), -1 (somewhat disagree), -2 (strongly disagree). 

\subsection{Nugget Type Annotations}\label{ss:nugget-annotation}

In Nugget Detection (ND) annotations, annotators were asked to identify nuggets for each dialogue, where a nugget is an turn that helps the Customer
transition from the current state (where the
problem is yet to be solved) towards the target
state (where the problem has been solved). \autoref{f:states} reflects our view that accumulating nuggets will eventually solve Customer's problem. 
The official definition of nuggets is 
\begin{enumerate}

\item A nugget is a turn by either Helpdesk or Customer;
\item It can neither partially nor wholly overlap with another nugget;
\item It helps Customer transition from Current State (including Initial State) towards Target State (i.e., when the problem is solved).
\end{enumerate}

Compared to the traditional nugget-based information access evaluation approaches,
there are two unique features in nugget-based  helpdesk dialogue evaluation:
\begin{itemize}[leftmargin=*, noitemsep,]
\item A dialogue involves two parties, Customer and Helpdesk;
\item Even within the same utterer,
nuggets are not homogeneous, by which we mean that some nuggets may play special roles.
In particular, since the dialogues we consider are task-oriented (but not {\em closed-domain},
which makes slot filling approaches infeasible),
there must be some nuggets that represent the state of {\em identifying} the task
and those that represent the state of {\em accomplishing} it.
\end{itemize}

Based on the above considerations, we defined the following four mutually exclusive nugget {\em types}:
\begin{description}[leftmargin=5em]
\item[CNUG0] Customer's {\em trigger nuggets}. These are nuggets that define customer's initial problem, which directly caused customer to contact Helpdesk.
\item[HNUG] Helpdesk's {\em regular nuggets}. These are nuggets in helpdesk's turns that 
are useful from Customer's point of view.
\item[CNUG] Customer's {\em regular nuggets}. These are nuggets in customer's turns
that are useful from helpdesk's point of view.
\item[HNUG$\ast$] Helpdesk's {\em goal nuggets}. These are nuggets in helpdesk's turns
which provide the customer with a solution to the problem.
\item[CNUG$\ast$] Customer's {\em goal nuggets}. These are nuggets in customer's turns
which tell helpdesk that customer's problem has been solved.

\item[CNAN] Customer's {\em not a nugget}. It means that the current customer turn does not help towards problem solving.
\item[HNAN] Helpdesk's {\em not a nugget}. It means that the current helpdesk turn does not help towards problem solving.

\end{description}
Note that each nugget type may or may not be present in a dialogue, and multiple nuggets of the same type may be present in a dialogue.

\subsection{DCH-2 Annotation Statistics}

\begin{table*}[t]
\begin{center}

\caption{DCH-2 annotation statistics.
Each dialogue was annotated independently by either 19 or 20 annotators.}\label{t:a-statistics}

\begin{tabular}{l|l|l|l|l|l}
\hline
\multicolumn{6}{l}{(a) Total  number and ratio of dialogue quality labels over all 4,390 dialogues}\\
\hline
					&$-2$		&$-1$		&0		&1		&2\\
\hline		
Task accomplishment	& 13937 (16.649\%)& 15497 (18.513\%)& 33810 (40.389\%) & 13659 (16.317\%) & 6807 (8.132\%)\\
Customer satisfaction	& 12877 (15.383\%) & 14829 (17.715\%) & 36754 (43.906\%) & 13334 (15.929\%) & 5916 (7.067\%)\\
Dialogue effectiveness	& 12643 (15.103\%)& 12308 (14.703\%)& 24810 (29.638\%) & 25397 (30.339\%) & 8552 (10.216\%)\\
\hline
\multicolumn{6}{l}{(b) Total number and ratio of turn-level nugget type labels over all 4,390 dialogues}\\
\hline
					&Trigger		&Regular		&Goal	&\multicolumn{2}{c}{Not-a-Nugget}\\
\hline
Customer turns		& 71925 (37.017\%) & 71115 (36.600\%) & 8079 (4.158\%)& \multicolumn{2}{c}{43186 (22.226\%)}\\
Helpdesk turns		&N/A & 93542 (59.414\%)& 23557 (14.963\%) & \multicolumn{2}{c}{40341 (25.623\%)}\\
\hline
\end{tabular}
\end{center}
\end{table*}

\begin{table}[t]
\begin{center}

\caption{Inter-rater agreement of DCH-2. Measured by Krippendorff's $\alpha$ coefficient.\label{t:agreement}}

\begin{tabular}{l|r}
\hline		
Task accomplishment & 0.301\\
Customer satisfaction 	 	& 0.213\\
Dialogue effectiveness		 & 0.323\\
Nugget & 0.395\\
\hline

\end{tabular}
\end{center}
\end{table}

Table~\ref{t:a-statistics} shows the statistics of the dialogue-level and turn-level annotations provided in DCH-2. \autoref{t:agreement} shows the inter-annotator agreement of DCH-2 measured by Krippendorff's $\alpha$ coefficient. It can be observed that the agreement among the annotators 
is not high.
which reflects the highly subjective nature of this annotation task. 
Evaluating such a customer-helpdesk dialogue is even subjective and difficult for human, and often there is no such thing as the ground truth: different people may have different opinions about the dialogue.

{It should be stressed that our annotation task is not like document relevance assessments, and that it is inherently highly subjective. In a previous study \cite{zeng18}, the authors found that annotators usually have different interpretation about evaluating such complicated customer-helpdesk dialogues. For example, when a user encounter a software bug, the helpdesk may promise to fix this problem in the next version. Some annotators think it is acceptable, but the other do not. Thus, we believe that hiring many assessors and preserving their different viewpoints in the data set, is more important than trying to force them into reaching an agreement.} We hired 19 or 20 annotators for each dialogue, so that the distributions that may preserve different viewpoints are formed by the annotations. For example, the customer satisfaction annotation of a dialogue may be  ``15 of 20 dialogues strongly agree that the customer was satisfied by the dialogue, but the other annotators somewhat agree it''. 

\section{Utilising DCH-2}\label{s:tool}
\subsection{Distribution}
To obtain DCH-2 from us for research purpose use, please contact the authors and sign a user agreement form first.

\subsection{Task}
To utilise DCH-2, researchers may follow the task setting in NTCIR-15 DialEval-1~\cite{zeng20} to train and test estimators that can automatically evaluate customer-helpdesk dialogues. Since the dialogues have both nugget annotations and quality annotations, there are two subtasks available: Nugget Detection (ND) and Dialogue Quality (DQ). The ND subtask requires to automatically classify the nugget type for each dialogue turn, and the DQ subtask is to estimate the three quality scores based on the criteria introduced in Section~\ref{ss:subjective-annotation}.  Also, since DCH-2 is a fully parallel data set that contains both Chinese dialogues and English dialogues, it can be used for other purposes, such as exploring the difference between Chinese task-oriented dialogues and English task-oriented dialogues.
\subsection{Evaluation}
Instead of evaluating both ND and DQ subtasks as simple classification problems using metrics like accuracy, we recommend the researchers to estimate the probability distributions on all the quality scores or nugget types to incorporate the highly subjective nature of dialogue evaluation. For example, a customer satisfaction estimator may output something like {\em70\% strongly agree with 30\% somewhat agree}.
To evaluate the effectiveness of the estimators, the distance between the estimated distribution and the golden distribution formed by the annotators can be calculated. Specifically, for the DQ subtask (an ordinal quantification task), we recommend using Normalised Match Distance (NMD; a special case of the Earth Mover's Distance) and
the Root Symmetric Normalised Order-aware Divergence (RSNOD). For the ND subtask (a nominal quantification task), we recommend using 
Jensen-Shannon Divergence (JSD) and Root Normalised Sum of Squares (RNSS) ~\cite{sakai18sigir}. Also, if a researcher wants to use DCH-2 to evaluate DQ and/or ND tasks
under the NTCIR task settings,
they can utilise our Python evaluation script to compute the official evaluation measure scores. The evaluation tool will be distributed along with DCH-2 data set.

\section{Conclusions}\label{s:conclusions}

We have described DCH-2, a parallel-corpus of 4,390 customer-helpdesk dialogues
with dialogue-level dialogue quality annotations and turn-level nugget annotations.
The data set is available at \url{https://dialeval-2.github.io/DCH-2/}. We described how we constructed the test collection and the philosophy behind it.

For the upcoming NTCIR-16 DialEval-2 task,
DCH-2 will be used as the training data for participants.
However, as we have discussed in Section~\ref{s:intro},
it may be leveraged for other purposes.





\bibliographystyle{ACM-Reference-Format}


\balance

\bibliography{sigir21dch2}


\end{document}